\documentclass{article}
\usepackage{enumitem}
\usepackage{spconf,amsmath,graphicx}
\usepackage{graphicx}
\usepackage{amsmath}
\usepackage{multirow}
\usepackage{caption2}
\usepackage{booktabs}
\usepackage{amsfonts}
\usepackage{float}
\usepackage{lipsum}
\usepackage[pagebackref=true,breaklinks=true,letterpaper=true,colorlinks,bookmarks=false,citecolor=blue,linkcolor=blue]{hyperref}

\title{Three-Dimensional Convolutional Neural Network Pruning with Regularization-Based Method}
\name{Yuxin Zhang$^{1}$, Huan Wang$^{1}$, Yang Luo$^{1}$, Lu Yu$^{2}$, Haoji Hu$^{2,*}$, Hangguan Shan$^{2}$, Tony Q. S. Quek$^{3}$}
\address{$^{1}$College of Information Science and Electronic Engineering, Zhejiang University, China \\
         $^{2}$ SUTD-ZJU IDEA, Hangzhou, China \\
         $^{3}$ Information Systems Technology and Design Pillar, \\ Singapore University of Technology and Design, Singapore 
}
\begin{document}
%
\maketitle
\newcommand\blfootnote[1]{%
\begingroup
\renewcommand\thefootnote{}\footnote{#1}%
\addtocounter{footnote}{-1}%
\endgroup
}

\blfootnote{*Corresponding Author.}
\begin{abstract}
 Despite enjoying extensive applications in video analysis, three-dimensional convolutional neural networks (3D CNNs) are restricted by their massive computation and storage consumption.
 To solve this problem, we propose a three-dimensional regularization-based neural network pruning method to assign different regularization parameters to different weight groups based on their importance to the network.
 Further we analyze the redundancy and computation cost for each layer to determine the different pruning ratios.
 Experiments show that  pruning based on our method can lead to $2\times$ theoretical speedup with only $0.41\%$ accuracy loss for 3D-ResNet18 and $3.28\%$ accuracy loss for C3D.
 The proposed method performs favorably against other popular methods for model compression and acceleration.
\end{abstract}

\begin{keywords}
3D CNN, video analysis, model compression, structured pruning, regularization
\end{keywords}

\section{Introduction}
\label{sec:intro}
Recent years have witnessed great progress in computer vision tasks powered by convolutional neural networks (CNNs), such as classification~\cite{Simonyan2014Very,googlenet}, detection~\cite{fast-rcnn,yolo}, and segmentation~\cite{FCN,deeplab}.
Various pre-trained models can be used to extract image features.
However, due to the lack of motion modeling,  two-dimensional convolutional neural networks (2D CNNs) can not directly be applied to extract the features of videos.
Thus, researchers propose various three-dimensional convolutional network (3D CNN) architectures.
In~\cite{Xu20123D,Karpathy2014Large}, the authors use 3D CNN to identify human actions in videos.
Tran \emph{et al.} propose a generic 3D CNN for action recognition which contains 1.75 million parameters~\cite{Du2015Learning}.
However, the high dimensions of 3D CNNs leads to massive computation and storage consumption, hindering its deployment on mobile and embedded devices.

In order to reduce the computation cost, researchers propose various methods to compress CNN models, including knowledge distillation~\cite{Hin2015Knowledge}, parameter quantization~\cite{Courbariaux2015BinaryConnect,Rastegari2016XNOR}, matrix decomposition~\cite{Zhang2015Accelerating} and parameter pruning~\cite{Han2015Deep}.

Parameter pruning is a promising approach for CNN compression and acceleration.
One problem of parameter pruning is that it often produces unstructured and random connections which is hard to implement on hardware platforms~\cite{HanLiuMao16}.
Thus many works focus on structured pruning which can shrink a network into a thinner one so that the implementation of the pruned network is efficient~\cite{AnwSun16,SzeCheYanEme17}.

There are two categories for structured pruning: importance-based methods and regularization-based methods.
Existing regularization-based methods tend to use the same regularization parameter for all weight groups in the network.
In~\cite{Wen2015Learning,Lebedev2016Fast}, the authors use the same regularization parameter and adopt Group LASSO~\cite{Ming2006Model} for structured sparsity regularization.
Recently Wang \emph{et al.}~\cite{2018arXiv180409461W} use weight decay for structured sparsity regularization and vary the regularization parameters for different groups, pruning state-of-the-art CNN models with no loss of accuracy.
\begin{figure*}
    \vspace{-0.4cm} 
	\centering
	\includegraphics[width=16cm]{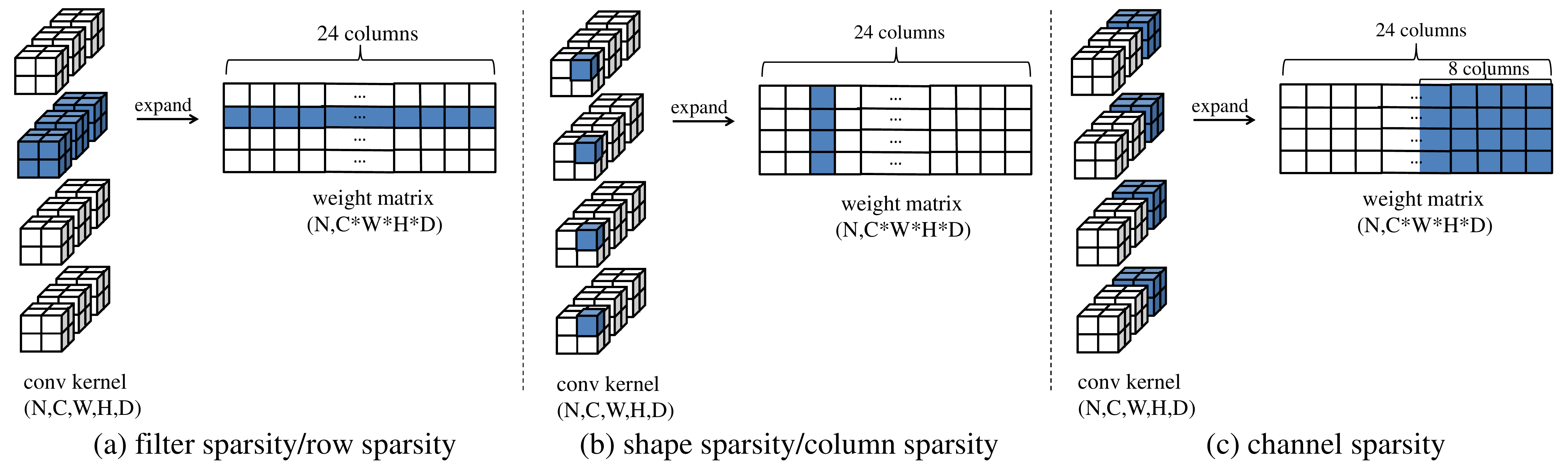}
	\caption{The im2col implementation~\cite{highp,ChetlurWVCTCS14} of 3D CNN is to expand tensors into matrices. Unlike 2D CNNs, each three-dimensional convolution kernel first expands its two-dimensional plane and then extends its depth. In this way convolutional operations are transformed to matrix multiplication. The weights at the blue squares are to be pruned.  (a) Pruning a filter. (b) Pruning all the weights at the same position. (c) Pruning a channel. In this paper, we adopt the shape sparsity.}
	\label{fig:structure}
\end{figure*}

However, all above methods focus on two-dimensional convolution, paying little attention to three-dimensional convolution.
In this paper, we use a regularization-based method to accelerate 3D CNN models.
The main idea is to add group regularization items to the objective function and prune weight groups gradually, where the regularization parameters for different weight groups are differently assigned according to some importance criterion.
Our contributions are threefold:
\begin{itemize}
\setlength{\itemsep}{0pt}
\setlength{\parsep}{0pt}
\setlength{\parskip}{0pt}
\item  An effective regularization-based pruning method is proposed to accelerate the 3D CNN models which receive relatively little attention compared with 2D CNNs in model compression and acceleration. 
\item We propose a novel scheme to determine the different pruning ratios for different layers by analyzing the redundancy and computation cost for each layer.
\item Adequate experiments on the UCF101 dataset with two main 3D CNN models yield better results than two popular methods in model compression and acceleration.
\end{itemize}

\section{The Proposed Method}
\label{sec:format}
\subsection{Determine the Pruning Weight Groups}
\label{sec:dpg}
For a three-dimensional convolutional neural network with~$L$ convolution layers, the weights of the~$l$th ($1 \leq l \leq L$) convolution layer~$\mathbf{W}^{(l)} \in \mathbb{R}^{N^l \times C^l \times W^l \times H^l \times D^l}$ is a sequence of 5-D tensor.
Here~$N^l$ is the number of filters, $C^l$ is the number of channels, $W^l$ is the width, $H^l$ is the height and $D^l$ is the depth.
The proposed objective function for structured sparsity regularization is defined by Equation (\ref{eqn:objective function}),
\begin{equation}
\label{eqn:objective function}
E(\mathbf{W}) = L(\mathbf{W}) + \frac{\lambda}{2}R(\mathbf{W}) + \sum_{l=1}^L \sum_{g=1}^{G^{(l)}} \frac{{\lambda_g}^{(l)}}{2} R({\mathbf{W}_g}^{(l)}),
\end{equation}
where~$L(\mathbf{W})$ is the loss on data, $R(\mathbf{W})$ is the non-structured regularization (weight decay in this paper), ${\lambda_g}^{(l)}$ is the regularization parameter of the layer $l$, $R({\mathbf{W}_g}^{(l)})$ is the structured sparsity regularization on each layer and $G^{(l)}$ is the number of weight groups in the layer $l$.
The learned structure is determined by the way of splitting weight groups~\cite{Wen2015Learning}.
Pruning of different weight groups for 3D CNN is shown in Figure \ref{fig:structure}.

In~\cite{2018arXiv180409461W},  Wang~\emph{et al.} theoretically prove that by increasing the regularization parameter~$\lambda_g$, the magnitude of weights tends to be minimized.
Our approach on 3D CNNs is build upon Wang's method. Specifically, we assign different~$\lambda_g$ for the weight groups based on their importance to the network.
Here, we use the $L_1$ norm as a criterion of importance.

For some specific layer, our goal is to prune $RN_g$  weight groups of it, where $R$ is the pruning ratio and $N_g$ is total number of weight groups in the layer.
In other words, we need to prune $RN_g$ weight groups which are ranked lowest in the layer.
We sort the weight groups in ascending order based on the $L_1$ norms.
In order to remove the oscillation of ranks during one  training iteration, we average the rank through training iterations to obtain the average rank $\overline{r}_{avg}$ in $N$ training iterations: $\overline{r}_{avg} = \frac{1}{N}\sum_{n=1}^{N} r_n$.

The final average rank~$\overline{r}$ is obtained by sorting~$\overline{r}_{avg}$ of different weight groups in ascending order, making its range from~$0$ to~$N_g-1$.
The update of $\lambda_g$ is determined by the following formula:
\begin{equation}
\lambda_g^{(new)} = \lambda_g^{(old)}+\Delta \lambda_g.
\label{eqn:changing lambda}
\end{equation}
Here $\Delta \lambda_g$ is the function of average rank $\overline{r}$, we follow the formula proposed by Wang~\cite{2018arXiv180409461W} as follows:
\begin{small}
	\begin{equation}
	\Delta\lambda_g(\overline{r}) = \left\{
	\begin{aligned}
	& -\frac{A}{RN_g}\overline{r}+A          &  \text{\emph{ if} }  \overline{r} \leq RN_g\\
	&-\frac{A}{N_g(1-R)-1}(\overline{r}-RN_g)     & \text{\emph{ if} } \overline{r} >  RN_g,
	\end{aligned}
	\right.
	\label{eqn:punish_function}
	\end{equation}
\end{small}
where~$A$ is a hyper-parameter which controls the speed of convergence.
According to Equation (\ref{eqn:punish_function}), we can see that~$\Delta \lambda_g$ is zero when $\overline{r} = RN_g$.
Since we aim at pruning $RN_g$ weight groups in the end, we need to increase the regularization parameters of the weight groups whose ranks are below $RN_g$ to further decrease their $L_1$ norms; and for those with greater $L_1$ norms and ranks above~$RN_g$, we need to decrease their regularization parameters to further increase their $L_1$ norms.
In this way, we can ensure that exactly $RN_g$ weight groups are pruned at the final stage of the algorithm.
When we obtain $\lambda_g^{(new)}$, the weights can be updated through back-propagation deduced from Equation (\ref{eqn:objective function}).
\subsection{Determine the Pruning Ratios}
Before pruning the network according to the method in Section~\ref{sec:dpg}, the foremost problem is to determine the pruning ratio for each convolution layer. 
In this work, we compared two strategies.
The first is to set the same pruning ratio (SPR) for each layer, which is commonly used in pruning~\cite{Wang2017Structured}.
The second, which is our proposed scheme, is to set different pruning ratios (DPR) for different layers according to the redundancy and computation cost.

Usually, the user gives a total pruning ratio, and we need to reasonably allocate the pruning ratio to each layer to meet the total pruning ratio.
On the one hand, different layers have different sensitivity to pruning, for which we adopt Principal Component Analysis (PCA)~\cite{Karlpearson1901LIII} to measure the redundancy of layers.
On the other hand, in terms of utility, there is little need to prune the layer with little computation cost, which introduces an analysis of GFLOPs.
\subsubsection{PCA: Consider the redundancy}
The redundancy of the convolution layer is measured by reconstruction error ($E_r$).
The reconstruction error of the weights of a convolution layer is obtained as follows: First expand the weights into a matrix ($N \times CWHD$).
Second, according to PCA, we can obtain the reconstructed vectors ($V_r$), and the reconstruction error is defined as $\label{RE}
   E_r=\frac{\left\|V_o-V_r\right\|^2}{\left\|V_o\right\|^2}$,
where $V_o$ is the original vector.

The smaller the reconstruction error, the higher the redundancy of the layer and thus we need to set a higher pruning ratio.
For a more comprehensive consideration, the remaining principle component ratio $k$ is uniformly distributed as $k \in \left\{0, 0.05, 0.10, ..., 0.90, 0.95, 1\right\}$.
After we obtain the $E_r$, the pruning proportion based on PCA analysis of the $l$th layer $\alpha_l$ is set to be inversely proportional to $E_r$.
\subsubsection{GFLOPs: Consider the utility}
The pruning proportion based on GFLOPs analysis of the $l$th layer $\beta_l$ is determined by simply normalizing GFLOPs of convolution layers $G_l$: $\beta_l=\frac{G_l}{\sum_{l=1}^{L} G_l}$.
\subsubsection{Combine PCA analysis and GFLOPs analysis}
The final pruning proportion of the $l$th layer $\gamma_l$ is determined by combining the analyses of PCA and GFLOPs, as follows:
$\label{eqn:combine two}
\gamma_l=(1-w_{gflops}) \times \alpha_l + w_{gflops} \times \beta_l$.
The default value of $w_{gflops}$ is 0.8, which is set empirically.

After obtaining the relative proportion, for a certain overall pruning ratio ($pr$), Equation (\ref{pr}) should be satisfied,
\begin{equation}
\label{pr}
v\times {\sum_{l=1}^{L} (G_l \times \gamma_l)}= pr \times {\sum_{l=1}^{L} G_l}.
\end{equation}
Thus we can obtain the value of $v$, and the specific pruning ratio of each layer is $v\times \gamma_l$.
\begin{figure}[ht]
	\centering
	\includegraphics[width=7cm]{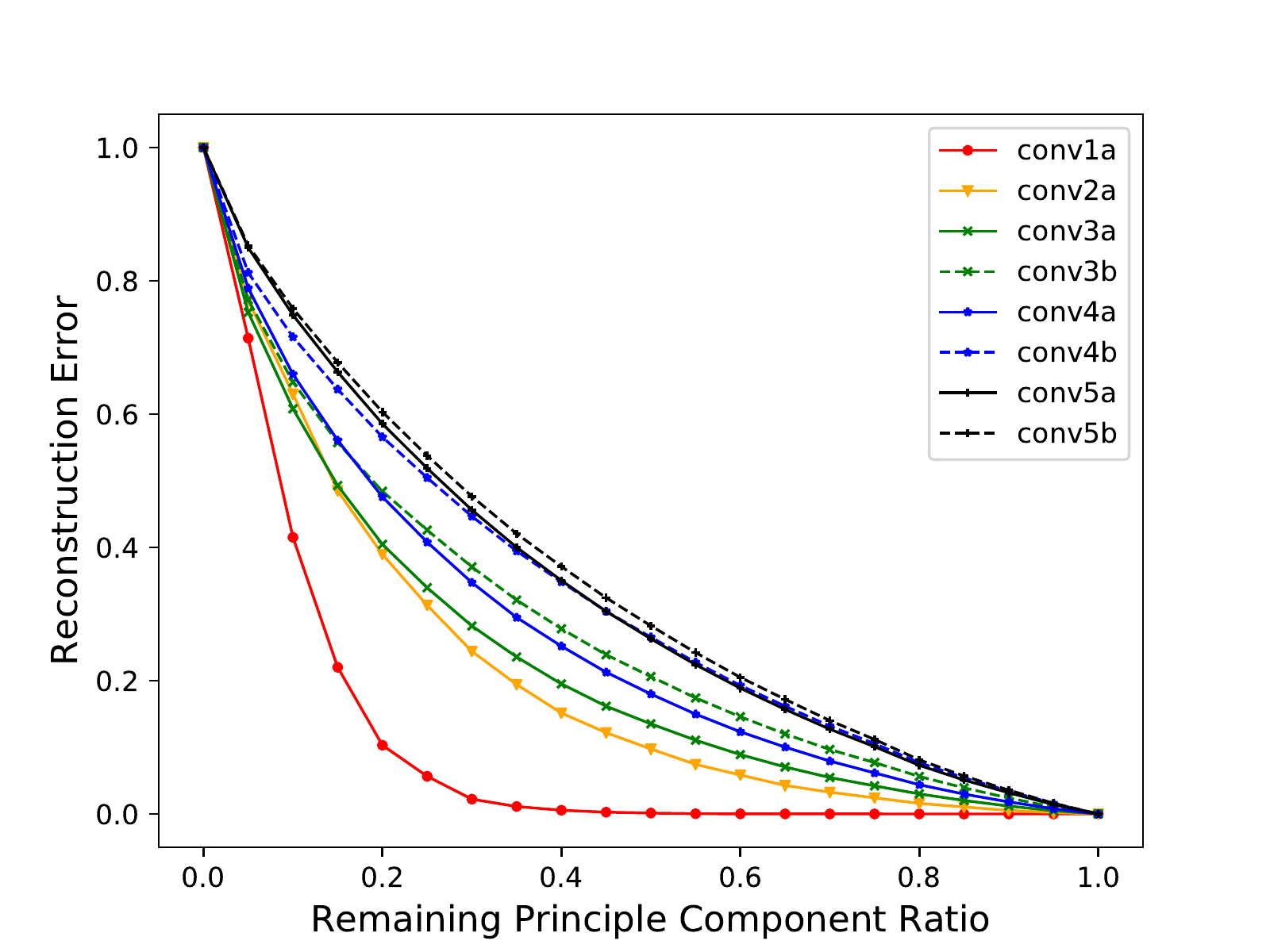}
	\caption{Reconstruction errors of different layers (C3D).}
	\label{fig:pca_c3d}
\end{figure}
\begin{figure}[htb]
	\centering
	\includegraphics[width=7cm]{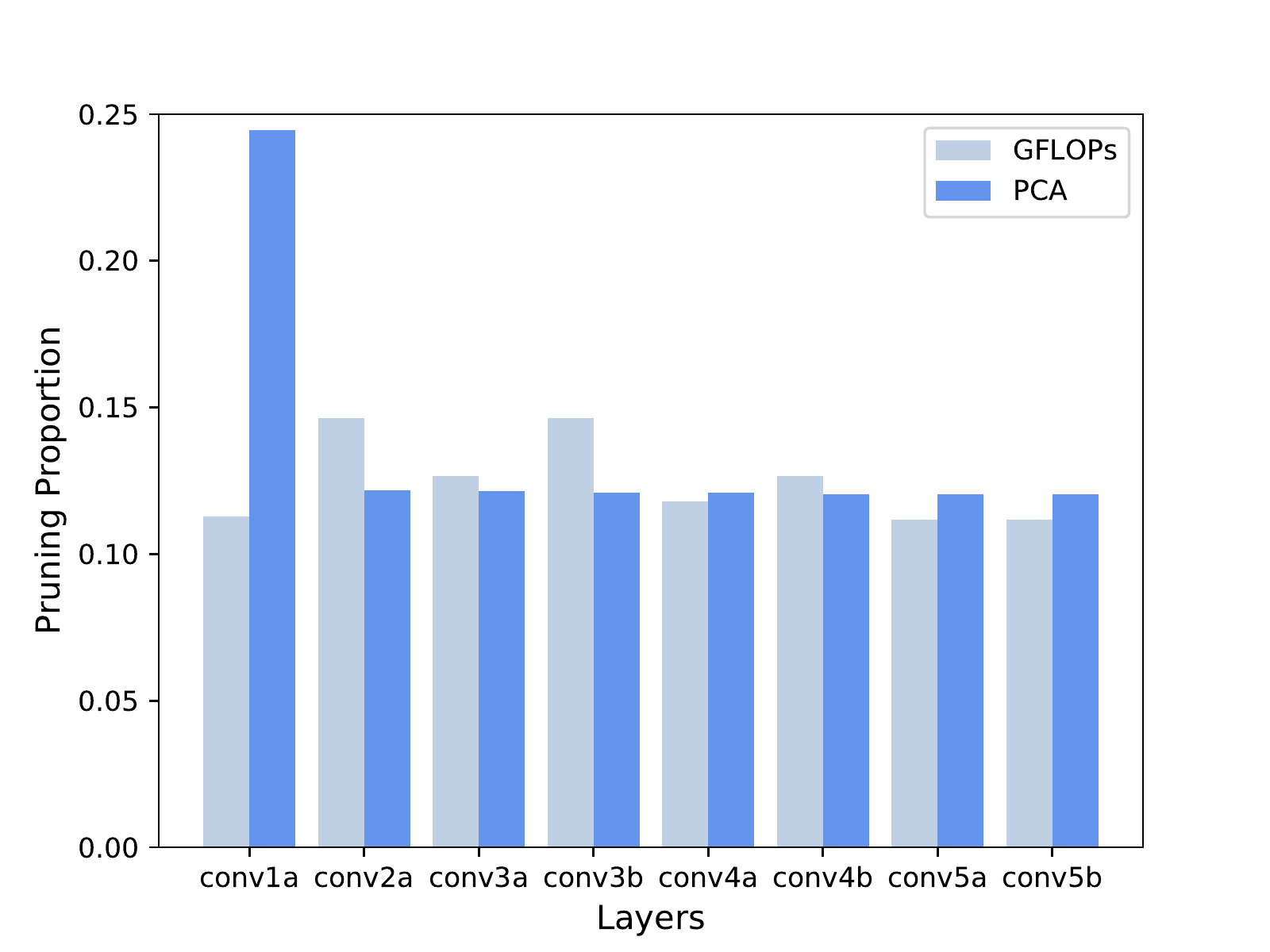}
	\caption{Normalized pruning proportion obtained by PCA analysis and GFLOPs analysis.}
	\label{fig:gp}
\end{figure}
\section{Experiments}
\label{sec:majhead}
Our experiments are carried out by Caffe~\cite{JiaSheDonEtAl14}.
We set the weight decay factor~$\lambda$ to be the same as the baseline and set hyper-parameter~$A$ to half of~$\lambda$.
We only compress the weights in  convolutional layers and leave the fully connected layers unchanged because we focus on network acceleration.
The methods used for comparison are Taylor Pruning (TP)~\cite{Molchanov2016Pruning} and Filter Pruning (FP) ~\cite{Li2016Pruning}.
For all experiments, the ratio of speedup is computed by GFLOPs reduction.
\begin{table*}[htb]
	\caption{Pruning ratios for convolution layers in C3D.}
	\label{prc3d}
	\centering
	\begin{tabular}{ccccccccc}
	\toprule 
	 Pruning ratio& conv1a& conv2a& conv3a& conv3b& conv4a& conv4b& conv5a& conv5b\\
	\midrule 
	2$\times$& 0.5129& 0.5279& 0.4684& 0.5273& 0.4414& 0.4676& 0.4222& 0.4222\\
	4$\times$& 0.7694& 0.7918& 0.7026& 0.7909& 0.6621& 0.7015& 0.6333& 0.6333\\
	\bottomrule 		
    \end{tabular}
\end{table*}
\begin{table}[ht]
	\caption{The increased error when accelerating C3D on UCF101 (baseline: 79.94\%).}
	\label{C3D}
	\centering
	\begin{tabular}{lcc}
		\toprule
		\multirow{2}*{Method}  &  \multicolumn{2}{c}{Increased err. (\%)} \\
		\cline{2-3}
		& $2\times$ & $4\times$\\
		\midrule
		TP~(our impl.)   &  $11.50$ & $21.19$  \\
		FP~(our impl.)   &  $4.92$ & $10.96$  \\
		Ours~(SPR) & $3.56$ & $7.02$ \\
		Ours~(DPR) & $\mathbf{3.28}$ & $\mathbf{6.56}$ \\	
		\bottomrule
	\end{tabular}
\end{table}
\begin{table}[ht]
	\caption{The increased error when accelerating 3D-ResNet18 on UCF101 (baseline: $72.50\%$).}
	\label{Res}
	\centering
	\begin{tabular}{lcc}
		\toprule
		\multirow{2}*{Method}  &  \multicolumn{2}{c}{Increased err. (\%)} \\
		\cline{2-3}
		& $2\times$ & $4\times$\\
		\midrule
		TP~(our impl.)   &  $5.72$ & $14.24$  \\
		FP~(our impl.)   &  $1.60$ & $6.92$  \\
		Ours~(SPR) & $0.91$ & $3.50$ \\
		Ours~(DPR) & $\mathbf{0.41 }$ & $\mathbf{2.87 }$ \\			
		\bottomrule
	\end{tabular}		
\end{table}
\subsection{C3D on UCF101}
\label{sec:c3d}
We apply the proposed method to C3D~\cite{Du2015Learning}, which is composed of $8$ convolution layers, $5$ max-pooling layers, and $2$ fully connected layers.
We download the open Caffe model as our pre-trained model, whose accuracy on UCF101 dataset is $79.94\%$.
UCF101 contains 101 types of actions and a total of $13320$ videos with a resolution of $320\times240$.
All videos are decoded into image files with $25$ fps rate.
Frames are resized into $128\times171$ and randomly cropped to $112\times112$.
Then frames are split into non-overlapped 16-frame clips which are then used as input to the networks. 
For all three methods, the learning rate is set to $0.0001$ and batch size is set to $30$.

In DPR, the reconstruction errors with different remaining principle component ratios are shown in the Figure \ref{fig:pca_c3d}.
It can be seen that, in the network of C3D, under the same remaining principle component ratio, the reconstruction error of upper layers (like conv5b) are greater than that of the bottom layers (like conv1a), which means that the upper layers are less redundant.
Figure \ref{fig:gp} shows the normalized pruning proportion obtained by PCA and GFLOPs analysis in DPR.
The final pruning ratios of each layer are shown in Table \ref{prc3d}.

The increased error caused by pruning is shown in Table \ref{C3D}.
With different speedup ratios, our approach is consistently better than TP and FP.
%
In addition, we can see that DPR outperforms SPR in both $2\times$ speedup and $4\times$ speedup.
In particular, DPR is especially effective when it achieves larger speedup ratio, e.g., $4\times$ speedup.
This further shows that it is very important to allocate the pruning ratio reasonably to each layer when pruning more parameters.
\begin{figure}[ht]
	\centering
	\includegraphics[width=7cm]{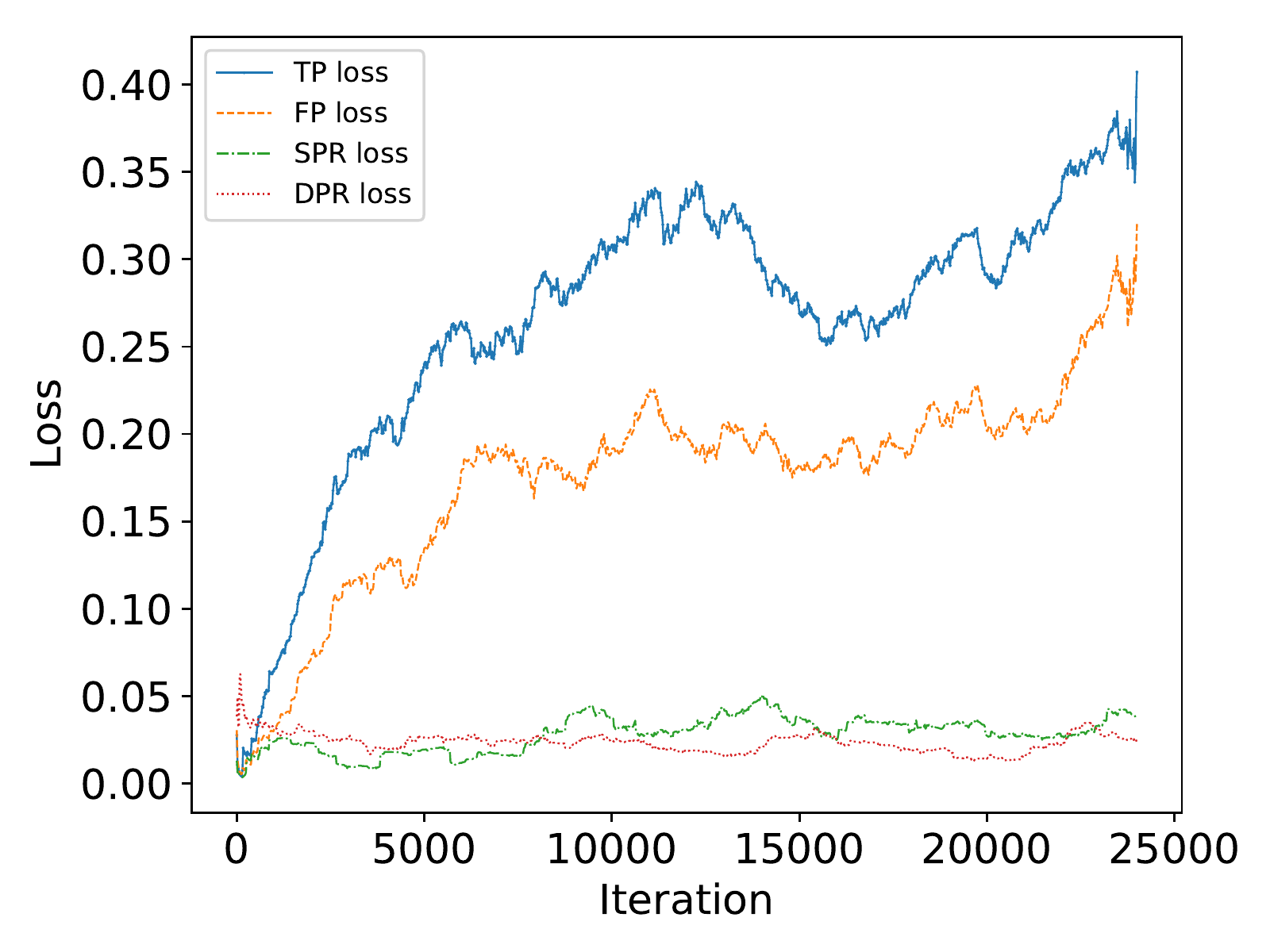}
	\caption{Training losses on 3D-ResNet18 for TP, FP and the proposed method.}
	\label{fig:all}
\end{figure}
\subsection{3D-ResNet18 on UCF101}
We further demonstrate our method on 3D-ResNet18~\cite{Du2015Learning}, which has $17$ convolution layers and $1$ fully-connected layer.
The network is initially trained on the Sport-1M database.
We download the model and then fine-tune it by UCF101 for $30000$ iterations, obtaining the accuracy of $72.50\%$.
The video preprocessing method is the same as stated in Section \ref{sec:c3d}.
The training settings are similar to that of C3D.

Experimental results are shown in Table  \ref{Res}.
The DPR only suffers $0.41\%$ increased error while achieving $2\times$ acceleration, obtaining better results than TP, FP, and SPR.
At the meantime, SPR also performs better than TP and FP.

Figure \ref{fig:all} shows the loss during the pruning process ($2\times$) for different methods.
As the number of iterations increases, the losses of TP and FP change dramatically, while the loss of our method remains at a lower level consistently.
This is probably because the proposed method imposes gradual regularization, making the network change little by little in the parameter space, while both the TP and FP direct prune less important weights once for all.
\section{Conclusion}
\label{sec:print}
In this paper, we propose a regularization-based method for 3D CNN acceleration.
By assigning different regularization parameters to different weight groups according to the importance estimation, we gradually prune weight groups in the network. 
The proposed method achieves better performance than other two popular methods in model compression.

\section{Acknowledgments}
This work is supported by the Natural Key R$\&$D Program of China under Grant 2017YFB1002400, the National Natural Science Foundation of China under Grant 61771427, the Natural Science Foundation of Zhejiang Province under Grant LY16F010004, the Zhejiang Provincial Public Technology Research of China under Grant 2016C31063, and the SUTD-ZJU IDEA Innovation Design Project (for Visiting Professor) under Grant 188170-11102/017 and 201804. 


\bibliographystyle{IEEEbib}
\bibliography{reference1}

\end{document}